\documentclass{article}

\usepackage{arxiv}

\usepackage[utf8]{inputenc} 
\usepackage[T1]{fontenc}    
\usepackage{hyperref}       
\usepackage{url}            
\usepackage{booktabs}       
\usepackage{amsfonts}       
\usepackage{nicefrac}       
\usepackage{microtype}      
\usepackage{lipsum}
\usepackage{graphicx}
\graphicspath{ {./images/} }
\usepackage{float}
\usepackage{caption}

\title{Analysis on DeepLabV3+ Performance for Automatic Steel Defects Detection}

\author{
Zheng Nie \\
  Stanford University\\
  \texttt{zhenie1988@gmail.edu} \\
   \And
 Jianchen Xu \\
  S\&P Global\\
  \texttt{jiachen.xu@spglobal.com} \\
  \And
 Shengchang Zhang \\
  University of Tennessee, Knoxville\\
  \texttt{sczhang63@utk.edu} \\
}

\begin{document}
\maketitle
\begin{abstract}
Our works experimented DeepLabV3+ with different backbones on a large volume of steel images aiming to automatically detect different types of steel defects. Our methods applied random weighted augmentation to balance different defects types in the training set. And then applied DeeplabV3+ model three different backbones, ResNet, DenseNet and EfficientNet, on segmenting defection regions on the steel images. Based on experiments, we found that applying ResNet101 or EfficientNet as backbones could reach the best IoU scores on the test set, which is around 0.57, comparing with 0.325 for using DenseNet. Also, DeepLabV3+ model with ResNet101 as backbone has the fewest training time. 

\end{abstract}


\section{Introduction}
Automatic defect detection based on steel surface images is a challenging task due to its diversity and complexity in the real industry, especially in the high-speed production lines, where the demand for real-time detection is essential but difficult for humans \cite{realtime}. Different environments' characteristics could affect the appearance of steel and its damage type. Existing methods could be coarsely divided into three main steps: image prepossessing, feature extraction, and classification. However, most of the times, the requirement of high-quality feature extractors are based on both hand-crafted works and adequate expert knowledge. \\

Because defect detection is based on images of steel surface, we conversed this problem to a image segmentation problem and applied the DeepLabV3+ model on solving it. DeepLabV3+ could be combined with different pre-trained models as backbones. To balance the accuracy and efficiency of the model, we experimented different backbones and have the detailed analysis of their performance.

\section{Related Works}
\label{sec:headings}
Among all architectures / tools applied for image segmentation, spatial pyramid pooling and encoder-decoder structure are two commonly applied tools.
Encoder-decoder structure applies encoder to gradually decrease feature maps and captures higher semantic information, decoder part to gradually recovers the spatial information. Well-studied models could be applied as the encoder module. In late 2014, Long et proposed 'Fully convolution network'\cite{FullyConv}, which utilized the well-studied image classification networks (eg. AlexNet) as encoder, appended a decoder module to upsample the coarse feature maps to the same size as the input image. In 2015, the U-Net \cite{U-Net} has been proposed, which consists of a contracting path for capturing context information for the image and a symmetric expanding path for precise localization.  \\
Spatial pyramid pooling works by keeping the partitioning image into smaller sub-regions and then taking a weighted sum of the number of matches at each sub-region level. \cite{SpatialPyramid}  DeepLabV3+, which reconstructed DeepLab architecture, has applied Atrous Spatial Pyramid Pooling (ASPP). With ASPP, the architecture could learn larger field-of-view without increasing the number of parameters or the amount of computation. And the output feature could have a larger size, which benefits to segmentation task. 
\section{Data}
\subsection{Overview}\vspace{-1em}
Data is from the Kaggle Competition - 'Severstal: Steel Defect Detection', a brunch of steel surface images. The objective is to automatically detect the region of defects of the steel. In each image, the steel could contain 0, 1, 2, 3 or all 4 kinds of defects. \\
The training set includes 12568 images. Among all these images, 6666 of them include at least one defect region. The ground truth of these images has been given by human labels. \\
The training set includes 7095 defect regions in total, but the defection classes are quite unbalanced in the aspect of the class type and defection area. To be more specific, around 73\% of defections are from class 3. Class 4 only maintain 11.3\% among all defections. But if consider the area, it is around 17\% among all defections, which means that this kind of defections is tended to be larger. On the contrary, the defect size for class 2 and class 1 is quite small. Although around 12.6\% and 3.48\% of defects come from these two classes, they only make up to 2.39\% and 0.51\% of the total mask respectively. In terms of sample and especially in terms of area,
 Our network may have a hard time finding classes 1 and 2 two because of their small size.
 
 \subsection{Data Preprocessing}
\begin{figure}[H]
    \centering
    \includegraphics[width=7cm, height=3cm]{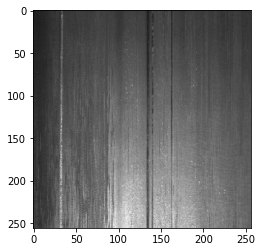}
    \includegraphics[width=7cm, height=3cm]{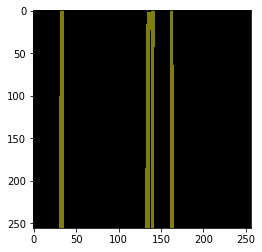}
    \caption{Resized image(left) and ground truth mask(right)}
    \label{fig:sample_rawdata}
\end{figure}

\begin{enumerate}
    \item Resize: reduced image size from 256 $\times$ 1700 to 256 $\times$ 256[Figure \ref{fig:sample_rawdata} (left)] for training. 
    \item Encoded segmentation and classification information as 2D numeric labeled masks matrix with same size as input data: Each element in the mask has value from 0 to 4. Background area is marked as 0. 1 to 4 corresponds to 4 types of defects. Visualize as Figure \ref{fig:sample_rawdata}( right).
\end{enumerate}

\subsection{Data Augmentation}
Applied random weighted augmentation\cite{cubuk2019randaugment} on training data to balance the defections. The logic is first calculating the proportion of each defection classes $P = [p_1, p_2, p_3, p_4]$ and then use $1 - P$ as the possibility of performing augmentation for each defect type. The augmentation is choosing randomly from the following types:
\begin{itemize}
    \item Random Crop: Randomly crop the image with the target input size from the raw image. 
    \item Vertical Flip: Flip the image vertically.
    \item Random Rotation: Rotate the image with a random angle.
\end{itemize}
The augmentation applies to both the original image and the corresponding ground truth masks to pair with the augmented images. \\
We add a weight(probability) $W = [w_1, w_2, w_3]$ for each type of augmentation during random selection for the augmentation actions.

\section{Evaluation Metrics}
In this project, we calculated IoU\cite{Rezatofighi_2019_CVPR} for each sample in the test data set, then get the mean value as the evaluation metrics. If $X_i$ denotes the input image, $Y_i$ denotes the mask matrix. N denotes total test data set size, mean IOU could be calculated as:
\[
mIoU = \frac{1}{N}\sum_i^N \frac{X_i \cap Y_i}{X_i \cup Y_i}
\]

\section{Models}
\subsection{\textbf{Baseline}}
\begin{figure}[H]
    \centering
    \includegraphics[scale=0.8]{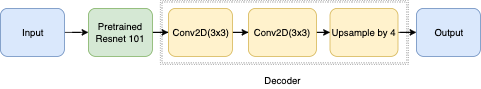}
    \caption{Bare Bone DeepLabV3+}
    \label{fig:baseline}
\end{figure}

The baseline model[Figure \ref{fig:baseline}] is a bare bone DeeplabV3+ by removing the encoder module. To be more specific, the input image will be fed into a pretrained Reset101 model then the DeeplabV3+ decoder module. The decoder module consists of 2 convolution layers and an up sample layer. Batch normalization and ReLu activation functions are added followed by the echo of these 2 convolution layers. The first convolution layer in the decoder module has 2048 input channels, 256 output channels, with kernel size 3. The second convolution layer has 256 input channels, 256 output channels, with kernel size 3. we kept the upsample layer to be consistent with DeepLabV3+ decoder structure.

\subsection{Experimented Model}
The experimented model is based on the DeepLabv3+ \cite{DeepLabV3p}, the overall architecture is shown below. 
\begin{figure}[htbp]  
    \centering 
    \includegraphics[width = 13cm, height = 5cm]{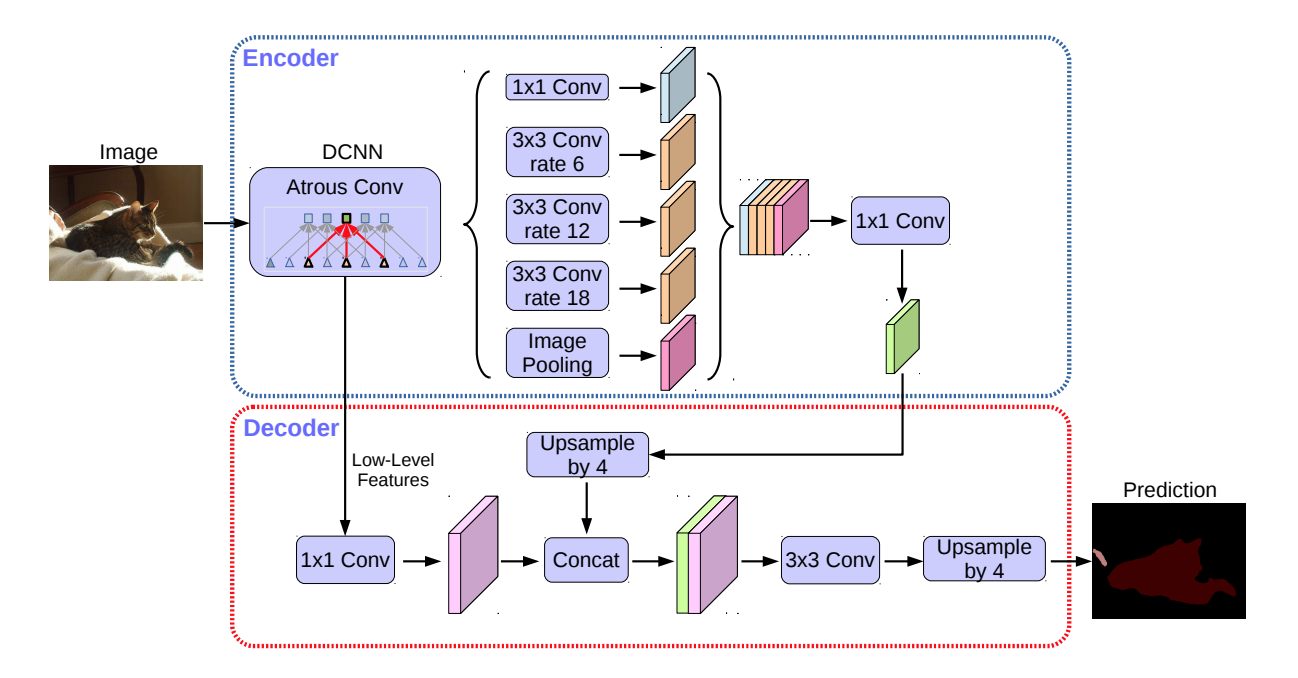}
\end{figure}\\

As the graph is shown, the model utilized the encoder-decoder structure. The encoder starts with 5 normal convolution modules, the output will be passed to 4 Atrous convolution modules and 1 average max-pooling module in parallel. The first 5 normal convolution modules are based on pre-trained backbone model, which describes as below subsections \ref{res_net}, the convolution layer has 7$\times$ 7,1$\times$1,1$\times$1,1$\times$1,1$\times$1 kernels, 2,1,2,2,1 strides, 0,1,1,1,2 dilation separately. The output of this module, on one hand, will be copied and saved as an input to the decoder to maintain more lower-level information. On the other hand, it will be passed to atrous convolution modules in parallel as the plot shown. The 4 atrous convolution modules have different convolution layer. The kernel of them are 1$\times$1,3$\times$3,3$\times$3,3$\times$3 and dilation values of them are 1, 12, 24, 3. The stride value is always 1. The average max-pooling module is a combination of an adaptive average Pooling layer and a normal convolution module to adjust the output size the same as the atrous convolution module.\\ The output from these 5 modules will be concatenated and passed to another depthwise convolution module aimed to merge information from different channels. Before the output pass to the decoder it will be up-sampled by 4 with bilinear methods.\\
The decoder part has two inputs, the outputs from the first 5 normal convolution modules and the outputs from the encoder part. The outputs from normal convolution modules are passed to a normal convolution module and then concatenates with the outputs from the encoder part. Finally, the concatenated outputs are passed to 3 other convolution modules. The kernel size of the convolution layer is 3$\times$ 3,3$\times$3 and 1$\times$ 1. The last step before the final output is up-sample the output by 4 with bilinear methods to make it has the same size as the input. 

\subsection{Backbone} \label{backbone}
Like we mentioned previously, We explored different pretrained models as backbone for DeeplabV3+, the backbone we experimented are:
\begin{itemize}
    \item ResNet\cite{he2016deep}: a pre-trained 101 layers ResNet
    \item DenseNet\cite{densenet}: a pre-trained 201 layers DenseNet
    \item EfficientNet\cite{efficientnet}: a pre-trained EfficientNet B1 model
\end{itemize}

\section{Experiments}
We implemented the model in PyTorch and project code in python, using Google Colab as the coding environment. But we there`s limited GPU available hours per day, and the log data may lose in between resume training. Due to the limitation of this, we only show part of the loss and evaluation trend here.

\subsection{Baseline}
We trained base line model on Google Colab with Nvida Telsa P100-GPU with learning rate=0.01, batch size = 16 weight decay = 1e-4, and data split is train/evaluation/test=80/10/10(\%). It took about 3 hours to reach mIoU = 0.30 after 30 epochs.

\subsection{DeepLabV3+}
\begin{figure}[H]
    \centering
    \includegraphics{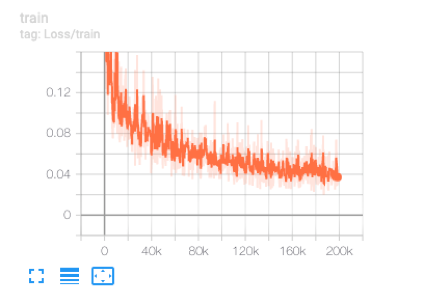}
    \includegraphics{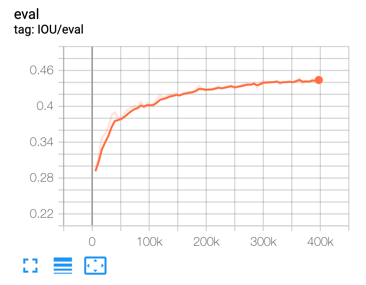}
    \caption{Loss(Left) and Mean IoU(right)}
    \label{fig:loss}
\end{figure}

We trained DeepLabV3+ for all backbones we explored in \ref{backbone} on Google Colab with Nvida Telsa P100-GPU with learning rate=0.01, batch size = 16 weight decay = 1e-4, and data split is train/evaluation/test=80/10/10(\%)Partial smoothed loss and evaluation trends shown as Figure [\ref{fig:loss}]. 

\subsection{Train/Evaluation Speed}
Total training time is based on trained iteration per seconds, the batch size is 16, so each iteration, the input is Batch x Channel x H x W matrix, which are 16 resized 256 x 256 images. \\
Evaluation time is based on image inferences per seconds, the batch size is set to 1.
\begin{center}
    \begin{table}[H]
    \begin{tabular}{c|c|c|c}
        \hline
        Model backbone & Train(iter/s) & Eval Resized Image(image/s) & Eval Original Image(image/s)  \\
        \hline
        Base & 24 & 26 & 11\\
        DenseNet201 & 25 & 10 & 7\\
        ResNet101 & 14 & 16 & 8\\
        EfficientNet B1 & 40 & 20 & 11 \\
        \hline
    \end{tabular}
    \caption{Speed Comparison}
    \label{tab:speed_compare}
    \end{table}
\end{center}
\section{Result}
\subsection{Mean IoU on Test Data}

\begin{table}[H]
\begin{center}
\begin{tabular}{ c|c } 
 \hline
 model & mIoU \\ 
 \hline
 Baseline & 0.29 \\ 
 DenseNet201 + DeeplabV3Plus & 0.325 \\
 ResNet101 + DeeplabV3Plus & 0.57 \\
 EfficientNet B1 + DeeplabV3Plus & 0.57 \\
 \hline
\end{tabular}
\caption{Mean IoU for different models}
    \label{tab:mIoU}
    \end{center}
\end{table}

\subsection{IoU Distribution on Test Data}
\begin{figure}[H]
    \centering
    \includegraphics[width=6cm, height=4cm]{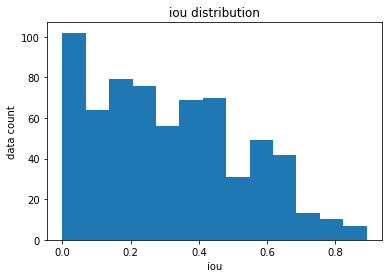}
    \includegraphics[width=6cm, height=4cm]{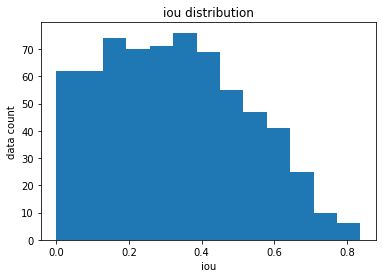}
    \includegraphics[width=6cm, height=4cm]{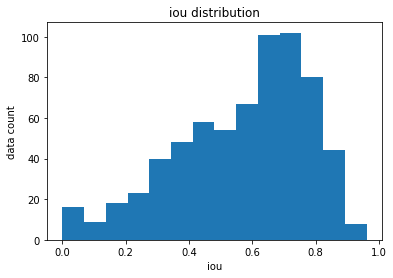}
    \includegraphics[width=6cm, height=4cm]{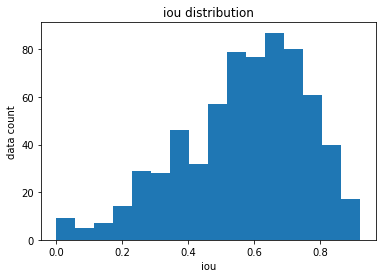}
    \caption{IoU Distribution: Baseline(First Row Left), DenseNet Backbone(First Row Right), EfficientNet B1 Backbone(Second Row Right) VS ResNet Backbone(Second Row Left)}
    \label{fig:iou_distro}
\end{figure}

IoU distribution statics for 4 models that experimented shown as [Figure \ref{fig:iou_distro}]
\subsection{Inference Sample on Test Data}
Perform model inference on test data with original image size(256 x 1700) and also compared with re-sized image(256 x 256).
\subsubsection{Single Defect Type per Image}
\begin{figure}[H]
        \centering
        \includegraphics[scale=0.9]{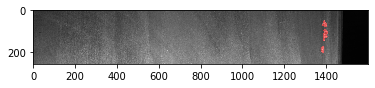}
        \caption*{a.) GroundTruth}
        \includegraphics[scale=0.9]{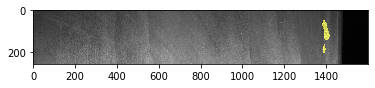}
        \caption*{b.) Base Model}
        \includegraphics[scale=0.9]{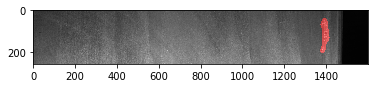}
        \caption*{c.) ResNet101 + DeeplabV3Plus}
        \includegraphics[scale=0.9]{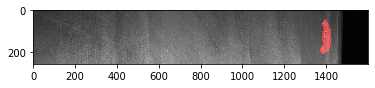}
        \caption*{c.) EfficientNetB1 + DeeplabV3Plus}
        \includegraphics[scale=0.9]{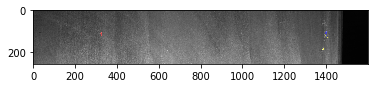}
        \caption*{d.) DenseNet201 + DeeplabV3Plus}
        \caption{Output comparison(resized)}
        \label{fig:Output_raw_size_single}
\end{figure}

[Figure \ref{fig:Output_raw_size_single}] shows output comparison between different models on original size images. The test data has only one defect type for one raw image. 

\subsubsection{Multiple Defect Type per Image}
\begin{figure}[H]
        \centering
        \includegraphics[scale=0.9]{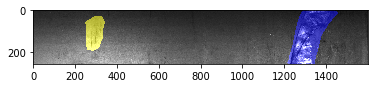}
        \caption*{a.) GroundTruth}
        \includegraphics[scale=0.9]{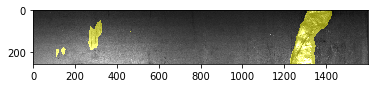}
        \caption*{b.) Base Model}
        \includegraphics[scale=0.9]{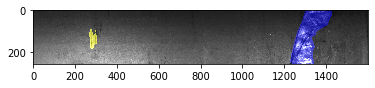}
        \caption*{c.) ResNet101 + DeeplabV3Plus}
        \includegraphics[scale=0.9]{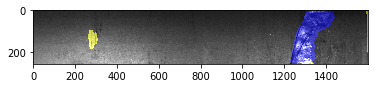}
        \caption*{c.) EfficientNetB1 + DeeplabV3Plus}
        \includegraphics[scale=0.9]{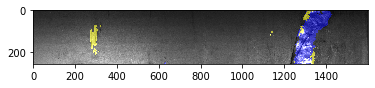}
        \caption*{d.) DenseNet201 + DeeplabV3Plus}
        \caption{Output comparison(resized)}
        \label{fig:Output_raw_size_mix}
\end{figure}

[Figure \ref{fig:Output_raw_size_mix}] shows output comparison between different models on original size images. The test data has multiple defect type for one raw image. 

\subsubsection{Resized Output}
\begin{figure}[H]
        \centering
        \includegraphics[scale=0.15]{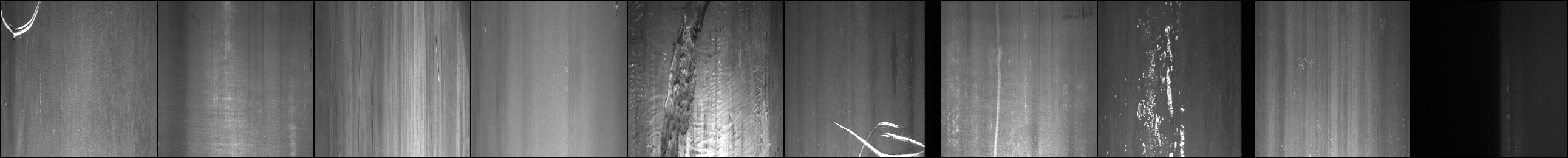}
        \caption*{a.) input image}

        \includegraphics[scale=0.15]{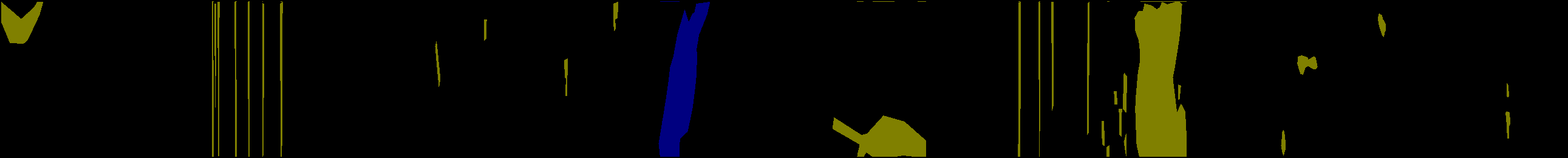}
        \caption*{b.) Ground truth label mask}

        \includegraphics[scale=0.15]{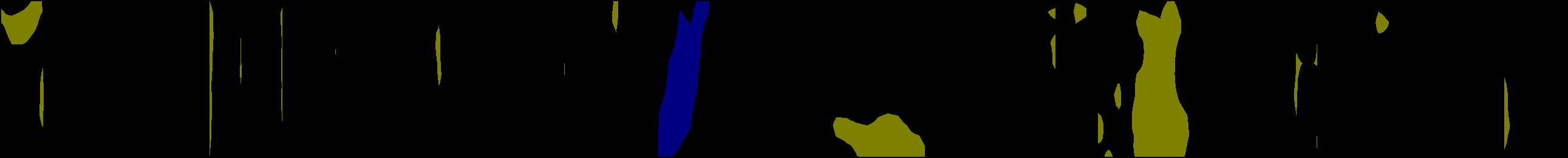}
        \caption*{c.) Base line output mask}
        
        \includegraphics[scale=0.15]{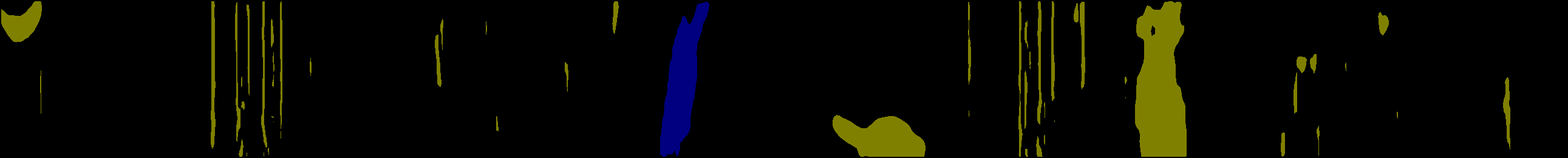}
        \caption*{d.) Resnet101 + DeeplabV3Plus output mask}
        \caption{Output comparison(resized)}
        \label{fig:Output}
\end{figure}

\section{Analyze} 
\begin{figure}[H]
        \centering
        \includegraphics[scale=0.9]{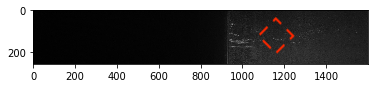}
        \caption*{a.) Raw Image}
        \includegraphics[scale=0.9]{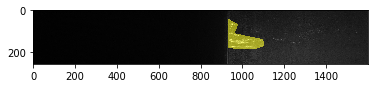}
        \caption*{b.) Groud Truth}
        \includegraphics[scale=0.9]{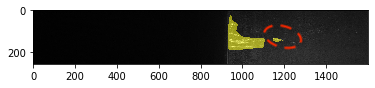}
        \caption*{c.) What Model See}
        \caption{Missing partial ground truth mask}
        \label{fig:output_missing_ground_truth}
\end{figure}

\begin{figure}[H]
        \centering
        \includegraphics[scale=0.9]{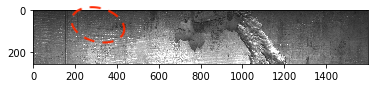}
        \caption*{a.) Raw Image}
        \includegraphics[scale=0.9]{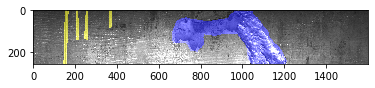}
        \caption*{b.) Groud Truth}
        \includegraphics[scale=0.9]{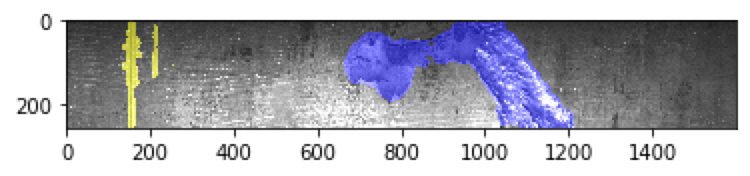}
        \caption*{c.) What Model See}
        \caption{Model limitation}
        \label{fig:too_small_too_see}
\end{figure}

Comparing the outputs from all four models with ground truth label[Figure \ref{fig:Output_raw_size_single}, \ref{fig:Output_raw_size_mix}, \ref{fig:Output}], we can see that the encoder module is helpful on learning details, like segmentation edges, and could make the detection region more precise. As a result of this, the IoU score could be improved with Encoder-Decoder architecture. When using  DeepLabV3+, ResNet and EfficientNet as the backbone, the IoU scores are pretty similar, when using EfficientNet as the backbone, the cost of training and evaluation is less. On the contrary, DenseNet does not perform well. \\ \\Some images have a relatively low IoU score [Figure \ref{fig:iou_distro}], which probably caused by the imbalanced data set and limitation of the model. With a deeper analysis on the underlying data set,  we found that the defections in 
\begin{itemize}
    \item Class 1 tends to have a smaller size and split in high fragments. It has the highest percentage of having more than 5 segments in one image but will occupy only a small total defection area. Also, it is the only class with considerable percentage of 10+ segment count.
    \item Class 2 tends to have a smaller size but not in many fragments. It doesn't have any 5+ segments per defect. Mostly comes in one or two segments.
    \item Class 3 has a fair amount of variation in terms of the number of segments and the area. 
    \item Class 4 is less frequent than in class 3 for having more than 3 segments per defect. 
\end{itemize}
based on these analysises, we thought that unbalanced data could lead to a low mIoU score.
\\ \\
By analyzing predictions with the low IoU score, we found some of the data were labeled improperly. However, the model could predict extra defects, which were not marked in the ground truth mask[Figure \ref{fig:output_missing_ground_truth}]. Some of the defect areas are detected by the model but not humans due to its extremely small size, which is invisible to human eyes[Figure \ref{fig:too_small_too_see}]. We also found some of the defects are too small to be labeled manually, but predicted by the model precisely. That could explain that some low the IoU score impacted overall the mean IoU score.

\section{Conclusion}
DeepLabV3+ with different backbones could be applied to detect and classify the steel defection automatically, both in fair accuracy and high efficiency. Among 4 experimented backbone, ResNet101 and EfficientNet have similar better performance, which IoU are around 0.57. In the meantime, ResNet101 is more efficient in training. The unbalanced data is a large difficulty in further increasing model performance. But the model could predict some extra regions, which too small to be recognized by humans but still potentially defected. We believe more and more explorations would be applied in this area to automatic more label-related tasks, which could be the part of the core idea of Industrial 4.0 \cite{ind4.0} manufacturing standard.

\section{Future Work}
We will continue improving the model mIoU performance and try to achieve a state-of-art result. We plan to try weighted loss\cite{weightedloss} and some other technique e.g. mixed-up\cite{Zhang2017mixupBE} to continue our experiment. We will try modifying the DeeplabV3+ to explore improvement on the model performance.



\begin{thebibliography}{10}

\bibitem{realtime}
Jiangyun Li, Zhenfeng Su, Jiahui Geng, and Yixin Yin.
\newblock Real-time detection of steel strip surface defects based on improved
  yolo detection network.
\newblock {\em IFAC-PapersOnLine}, 51:76--81, 01 2018.

\bibitem{FullyConv}
Jonathan Long, Evan Shelhamer, and Trevor Darrell.
\newblock Fully convolutional networks for semantic segmentation.
\newblock {\em CoRR}, abs/1411.4038, 2014.

\bibitem{U-Net}
Olaf Ronneberger, Philipp Fischer, and Thomas Brox.
\newblock U-net: Convolutional networks for biomedical image segmentation.
\newblock {\em CoRR}, abs/1505.04597, 2015.

\bibitem{SpatialPyramid}
Svetlana Lazebnik, Cordelia Schmid, and J.~Ponce.
\newblock Beyond bags of features: Spatial pyramid matching for recognizing
  natural scene categories.
\newblock volume~2, pages 2169 -- 2178, 02 2006.

\bibitem{cubuk2019randaugment}
Ekin~D. Cubuk, Barret Zoph, Jonathon Shlens, and Quoc~V. Le.
\newblock Randaugment: Practical automated data augmentation with a reduced
  search space, 2019.

\bibitem{Rezatofighi_2019_CVPR}
Hamid Rezatofighi, Nathan Tsoi, JunYoung Gwak, Amir Sadeghian, Ian Reid, and
  Silvio Savarese.
\newblock Generalized intersection over union: A metric and a loss for bounding
  box regression.
\newblock In {\em The IEEE Conference on Computer Vision and Pattern
  Recognition (CVPR)}, June 2019.

\bibitem{DeepLabV3p}
Liang{-}Chieh Chen, Yukun Zhu, George Papandreou, Florian Schroff, and Hartwig
  Adam.
\newblock Encoder-decoder with atrous separable convolution for semantic image
  segmentation.
\newblock {\em CoRR}, abs/1802.02611, 2018.

\bibitem{he2016deep}
Kaiming He, Xiangyu Zhang, Shaoqing Ren, and Jian Sun.
\newblock Deep residual learning for image recognition.
\newblock In {\em Proceedings of the IEEE conference on computer vision and
  pattern recognition}, pages 770--778, 2016.

\bibitem{densenet}
Gao Huang, Zhuang Liu, and Kilian~Q. Weinberger.
\newblock Densely connected convolutional networks.
\newblock {\em CoRR}, abs/1608.06993, 2016.

\bibitem{efficientnet}
Mingxing Tan and Quoc~V. Le.
\newblock Efficientnet: Rethinking model scaling for convolutional neural
  networks.
\newblock {\em CoRR}, abs/1905.11946, 2019.

\bibitem{ind4.0}
Andreja Rojko.
\newblock {Industry 4.0 Concept: Background and Overview}.
\newblock
  \url{https://online-journals.org/index.php/i-jim/article/viewFile/7072/4532},
  2017.

\bibitem{weightedloss}
Fereshte Khani, Aditi Raghunathan, and Percy Liang.
\newblock Maximum weighted loss discrepancy.
\newblock {\em CoRR}, abs/1906.03518, 2019.

\bibitem{Zhang2017mixupBE}
Hongyi Zhang, Moustapha Ciss{\'e}, Yann Dauphin, and David Lopez-Paz.
\newblock mixup: Beyond empirical risk minimization.
\newblock {\em ArXiv}, abs/1710.09412, 2017.

\end{thebibliography}

\end{document}